%% file: root.tex
\documentclass[letterpaper, 10 pt, conference]{IEEEtran}  

\IEEEoverridecommandlockouts                              

\usepackage{graphics} 
\usepackage{amsmath} 

\usepackage{graphicx}
\usepackage{caption}
\usepackage{subcaption}
\usepackage{float}
\usepackage{booktabs}
\usepackage{siunitx}
\usepackage{enumerate}
\usepackage{bm}
\usepackage{amssymb} 
\usepackage{pifont}  
\usepackage{overpic}
\usepackage{comment}
\usepackage{cuted} 
\usepackage{hyperref}
\usepackage[capitalise]{cleveref}
\usepackage{adjustbox} 

\usepackage{xcolor}
\newcommand{\xmark}{\ding{55}} 
\usepackage{eso-pic}

\newcommand\AtPageUpperCenterNotice[1]{%
  \AtPageUpperLeft{%
    \put(\LenToUnit{0.5\paperwidth},\LenToUnit{-2cm}){\makebox[0pt]{#1}}%
  }%
}

\AddToShipoutPictureBG*{%
  \AtPageUpperCenterNotice{%
    \parbox[b][2cm][c]{\paperwidth}{%
      \centering
      \fontsize{12}{14}\selectfont
      \color{gray!50}
      This paper has been accepted for publication at the\\
      Intelligent Vehicles Symposium (IV), Romania 2025. \copyright{}IEEE
    }%
  }%
}

\AddToShipoutPictureBG*{
  \AtPageLowerLeft{%
    \raisebox{25pt}{\makebox[\paperwidth]{\begin{minipage}{21cm}\centering
    \fontsize{10}{12}\selectfont
    \textcolor{gray!50}{%
      \copyright{} 2025 IEEE. Personal use of this material is permitted.
      Permission from IEEE must be obtained for all other uses, in any current or future media,
      including reprinting/republishing this material for advertising or promotional purposes,
      creating new collective works, for resale or redistribution to servers or lists,
      or reuse of any copyrighted component of this work in other works.
    }
    \end{minipage}}}%
  }
}

\title{\LARGE \bf
TinyCenterSpeed: Efficient Center-Based Object Detection for Autonomous Racing 
}

\author{Neil Reichlin\IEEEauthorrefmark{1}\IEEEauthorrefmark{2}, Nicolas Baumann\IEEEauthorrefmark{1}\IEEEauthorrefmark{2}\IEEEauthorrefmark{3}, Edoardo Ghignone\IEEEauthorrefmark{2}, Michele Magno\IEEEauthorrefmark{2}
\thanks{\IEEEauthorrefmark{1}Equal Contribution}%
\thanks{\IEEEauthorrefmark{2}Center for Project-based Learning, D-ITET, ETH Zurich}%
\thanks{\IEEEauthorrefmark{3}Corresponding author:
        {\tt\small nicolas.baumann@pbl.ee.ethz.ch}}%
}

\input{glossaries} 

\begin{document}

\maketitle

\begin{strip}
    \vspace{-2.0cm}
    \centering
    \includegraphics[angle=0,origin=c,width=\textwidth]{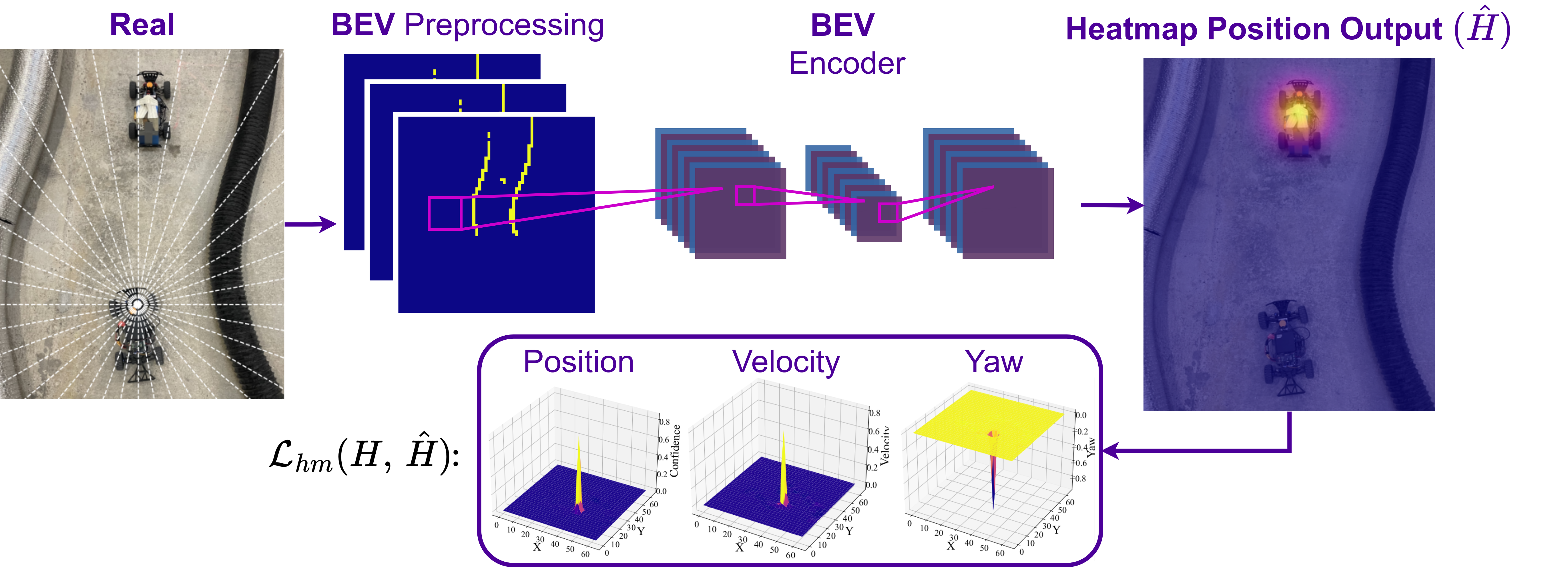}
    \captionof{figure}{Schematic depiction of the proposed \emph{TinyCenterSpeed} architecture to estimate the position, velocity, and heading of opponent autonomous racing cars. The heatmap loss $\mathcal{L}_{hm}$ is inspired from \emph{CenterPoint}, but the architecture is optimized for computationally constrained \glspl{obc} such as encountered in 1:10 scaled autonomous racing.}
    \label{fig:graphical_abs}
    \vspace{-0.5cm}
\end{strip}

\thispagestyle{empty}
\pagestyle{empty}

\begin{abstract}
    Perception within autonomous driving is nearly synonymous with \glspl{nn}. Yet, the domain of autonomous racing is often characterized by scaled, computationally limited robots used for cost-effectiveness and safety. For this reason, opponent detection and tracking systems typically resort to traditional computer vision techniques due to computational constraints. This paper introduces \emph{TinyCenterSpeed}, a streamlined adaptation of the seminal \emph{CenterPoint} method, optimized for real-time performance on 1:10 scale autonomous racing platforms. This adaptation is viable even on \glspl{obc} powered solely by \glspl{cpu}, as it incorporates the use of an external \gls{tpu}. We demonstrate that, compared to \gls{abd}, the current \gls{sota} in scaled autonomous racing, \emph{TinyCenterSpeed} not only improves detection and velocity estimation by up to 61.38\% but also supports multi-opponent detection and estimation. It achieves real-time performance with an inference time of just \SI{7.88}{\milli\second} on the \gls{tpu}, significantly reducing \gls{cpu} utilization 8.3-fold.
\end{abstract}

\section{Introduction}
In the domain of general autonomous driving, the role of perception is predominantly occupied by \glspl{nn}. These systems typically utilize complex models trained on extensive datasets such as \cite{caesar2020nuscenes, waymo, kitti}. The advantage of \glspl{nn} lies in their ability to implicitly learn a rich set of features, enabling effective generalization across complex scenarios. Within this domain, the \emph{CenterPoint} \cite{yin2021center} architecture has been particularly influential, enabling high performance on benchmarks like \emph{nuScenes} by identifying the positions of other traffic participants through a heatmap loss that models positions as Gaussian distributions centered on the detected objects.

Conversely, the adoption of \gls{nn}-based perception in autonomous racing is less prevalent, mainly due to the simpler environmental dynamics of racing settings, which are more controlled, and the computational constraints associated with racing. These constraints are crucial because autonomous racing aims to push the boundaries of autonomy algorithms, often prioritizing available computational resources elsewhere, thus sidelining the computationally expensive \gls{nn} inference. This is particularly true in scaled autonomous racing on platforms like the 1:10 scaled \emph{F1TENTH} \cite{okelly2019f1tenth}, which are favored in research for their cost-effectiveness and safety compared to full-scale models \cite{baumann2024forzaeth}.

This scenario poses a critical question: Are scaled autonomous platforms potentially missing out on performance enhancements that \glspl{nn} could provide? Addressing this, \emph{TinyCenterSpeed} is introduced as a computationally efficient adaptation of the \emph{CenterPoint} architecture \cite{yin2021center} tailored for scaled autonomous racing. Leveraging the open-source \emph{ForzaETH} racing stack \cite{baumann2024forzaeth}, this work compares \emph{TinyCenterSpeed} against the classical \gls{sota} \gls{abd} detector and \gls{ekf} tracker for single-opponent state estimation. Our results reveal that the proposed architecture, with only two consecutive 2D \gls{lidar} measurements, significantly outperforms the traditional system, while also enabling velocity estimation for multiple opponents. Furthermore, by quantizing and offloading the computational load to a USB-connected external \gls{tpu}, the model performs in real-time on the \gls{obc}, enhancing speed over the classical \gls{abd} and reducing \gls{cpu} utilization by 8.3-fold.
See \Cref{fig:graphical_abs} for a schematical representation of \emph{TinyCenterSpeed}.
The contributions of this work are summarized as follows:

\begin{enumerate}[I]
    \item \textbf{Architecture:} The \emph{TinyCenterSpeed} \gls{nn} architecture, inspired by \emph{CenterPoint}, demonstrates up to a 61.38\% improvement in detection and velocity estimation compared to traditional \gls{abd} algorithms.
    \item \textbf{Computational Efficiency:} We showcase the benefits of quantized \gls{nn} inference and computational offloading to an external \gls{tpu}. Fully quantizing the model to \texttt{INT8} enables \gls{nn} processing in just \SI{7.88}{\milli\second} and reduces \gls{cpu} utilization by 8.3-fold compared to \gls{cpu} inference of the full-precision \gls{nn}.
    \item \textbf{Multi-Opponent Capability:} The traditional \gls{abd} detector requires an \gls{ekf} tracker to estimate the velocity states of opponents. Moreover, to simplify data association, the current \gls{sota} from \cite{baumann2024forzaeth} is limited to tracking only a single dynamic opponent. In contrast, \emph{TinyCenterSpeed} inherently supports the detection and velocity estimation of multiple opponents simultaneously, eliminating the need for a tracker. It achieves a \gls{mate} of \SI{0.159}{\metre} and a \gls{mave} of \SI{0.318}{\metre\per\second}.
    \item \textbf{Open Source:} The autonomous racing dataset used to train this \gls{nn}, along with the training scripts, the model weights, and a \gls{gt} extraction \gls{gui} and algorithm are fully open-sourced for the autonomous racing community at: \href{github.com/ForzaETH/TinyCenterSpeed}{\url{https://github.com/ForzaETH/TinyCenterSpeed}}. An experimental demonstration can be found at \href{https://youtu.be/X8cuOlfszvI}{https://youtu.be/X8cuOlfszvI}.
\end{enumerate}

\section{Related Work}
Current \gls{sota} in object detection \cite{cai2023bevfusion4d} relies on high-end computational platforms requiring desktop-grade \glspl{gpu} and high-dimensional sensors, such multi-camera systems and  3D \glspl{lidar} such as those available in commonly used datasets \cite{caesar2020nuscenes, waymo, kitti}.
These methods are therefore unsuitable for resource-constrained applications, such as scaled autonomous racing, where low latency is paramount.
Using 2D \gls{lidar} can be a computationally cheaper alternative: the work in \cite{chen2021pseudoimage}, for example, shows that even without the commonly used 3D \gls{lidar}, robust obstacle detection can be achieved. 
The work in \cite{jia2022twoDvsthreeD} further compares 2D ranging sensors with 3D ones for the specific task of person detection, concluding that 2D \glspl{lidar} offer indeed a better tradeoff between performance and computational requirements, which is a desirable trait for embedded deployment.
However, these methods are not directly targeting scaled autonomous racing application scenarios, where a different set of requirements is needed. 
For instance, the detector presented in \cite{baumann2024forzaeth} describes a system where \gls{lidar} data is processed via the \gls{abd} method, which results in a computationally light algorithm. Furthermore, velocity estimates are also obtained via a \gls{kf}: this piece of information can be crucial in autonomous racing scenarios in the context of planning or overtaking \cite{baumann2024PSpliner}.
Other scaled racing algorithms utilize 2D \gls{lidar} \cite{hell2024lidar, zarrar2024tinylidarnet} as an input to a \gls{nn}, however these methods present an end-to-end solution that directly provides control inputs to the vehicle, rather than only a detection system.
This crucial difference can have some important downsides: firstly the whole system becomes more of a \emph{black-box}, which complicates, for example, understanding in case of failure cases; furthermore, a detection system can be modularily integrated with other \gls{sota} planning or control methods such as the one presented in \cite{baumann2024PSpliner}, differently from an end-to-end system. 
\emph{TinyCenterSpeed} addresses this gap in the field of scaled autonomous racing by building on top of previous work. 
Firstly, it uses the commonly used \gls{bev} representation, which was shown to be very efficient for point cloud enconding \cite{zhou2018voxelnet, lang2019pointpillars}, very effective for perception tasks \cite{li2024delving, zhao2024bev, cai2023bevfusion4d} and specifically apt in 3D object detection \cite{chen2017multi}.
Secondly, by building on top of the center-based paradigm presented with \emph{CenterNet} \cite{zhou2019objects} and extended to 3D tracking in \gls{bev} space in \cite{yin2021center}, the presented system is able to further provide opponent velocity estimates, which improves the overall opponent state estimation and therefore can benefit the system as a whole.
Finally, a modified version of the \emph{Heatmap Weighting Loss} \cite{heatmaploss} is used, in order to increase the significance of non-zero values at keypoint locations against the background.

\Cref{tab:rel_work} presents the related work in the area of 2D \gls{lidar} systems for scaled autonomous racing.

\begin{table}[ht]
    \centering
    \begin{adjustbox}{max width=\columnwidth}
    \begin{tabular}{lcccc}
        \toprule
        & \acrshort{ml}-based & 2D \acrshort{lidar} & Detection  & Resource \\ 
        & &  &  System &  Constrained \\ 
        
        \midrule
        \gls{sota} detection \cite{zhao2024bev, li2024delving} & \checkmark & \xmark & \checkmark & \xmark \\ 
        \emph{ForzaETH} Race Stack \cite{baumann2024forzaeth} & \xmark & \checkmark & \checkmark & \checkmark \\ 
        2D \acrshort{lidar} detection \cite{chen2021pseudoimage, jia2022twoDvsthreeD} & \checkmark & \checkmark & \checkmark & \xmark \\ 
        TinyLidarNet \cite{zarrar2024tinylidarnet} & \checkmark & \checkmark & \xmark & \checkmark \\ 
        \emph{\textbf{TinyCenterSpeed}} (\textbf{Ours}) & \checkmark & \checkmark & \checkmark & \checkmark \\ 
        \bottomrule
    \end{tabular}
    \end{adjustbox}
    \caption{Overview of related work in the context of 2D \acrshort{lidar} systems and scaled autonomous racing.}
    \label{tab:rel_work}
\end{table}

\section{Methodology}
\subsection{Neural Network Architecture}
\begin{figure*}[ht] 
    \centering
    \includegraphics[width=0.95\linewidth, trim=0cm 0cm 0cm 0.4cm, clip]{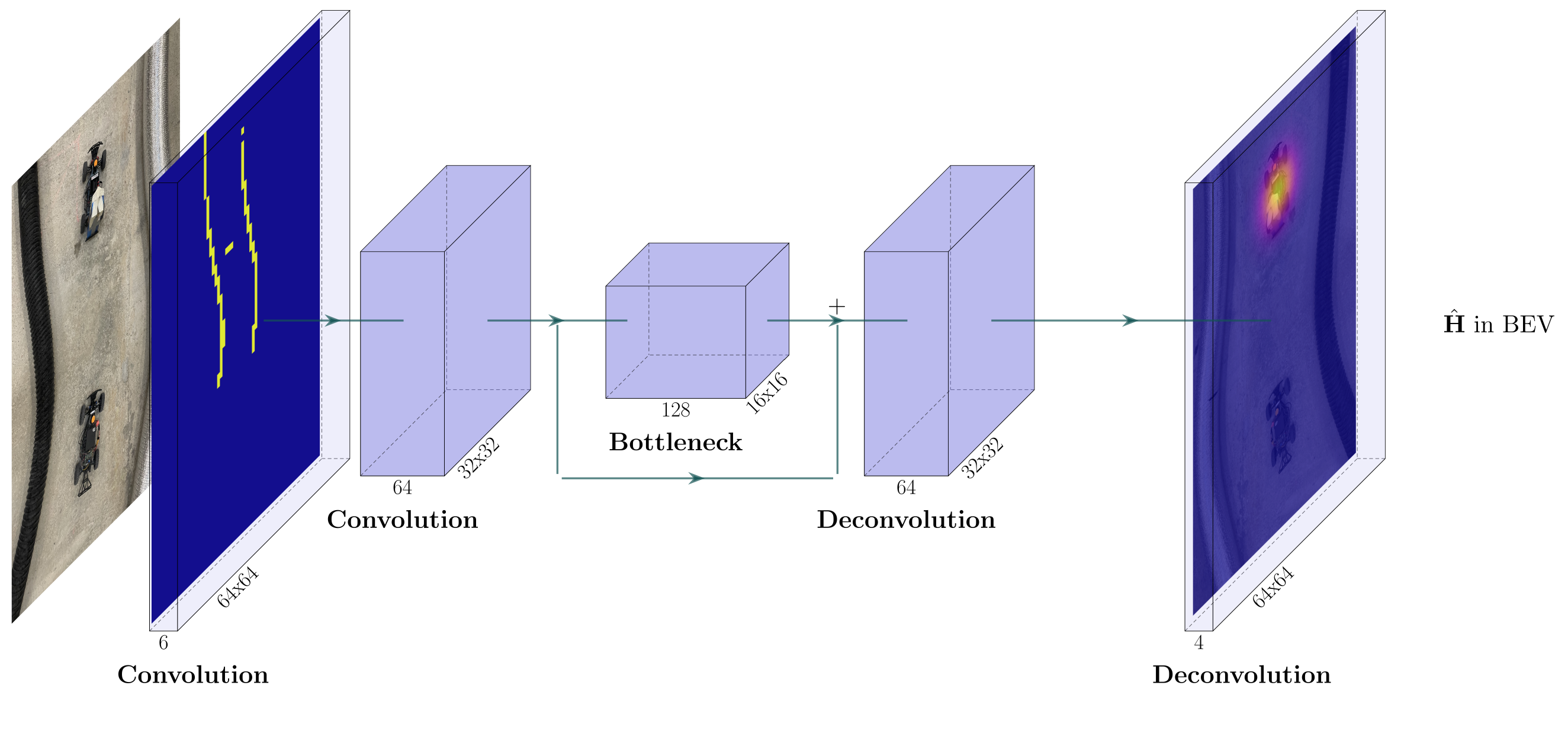}
    \vspace{-0.8cm}
    \caption{This figure exemplarily shows the architecture of \emph{TinyCenterSpeed}. The figure accurately depicts the model dimensions and shows the convolutional and deconvolutional layers as well as the residual connections used in the architecture. Note that the images of the robots are for visualization purposes only and are not used in the actual model.}
    \label{fig:CS-ARCH}
\end{figure*}

\begin{figure}[ht]
    \centering
    \includegraphics[width=0.8\linewidth, trim=0cm 0.4cm 0cm 0.2cm, clip]{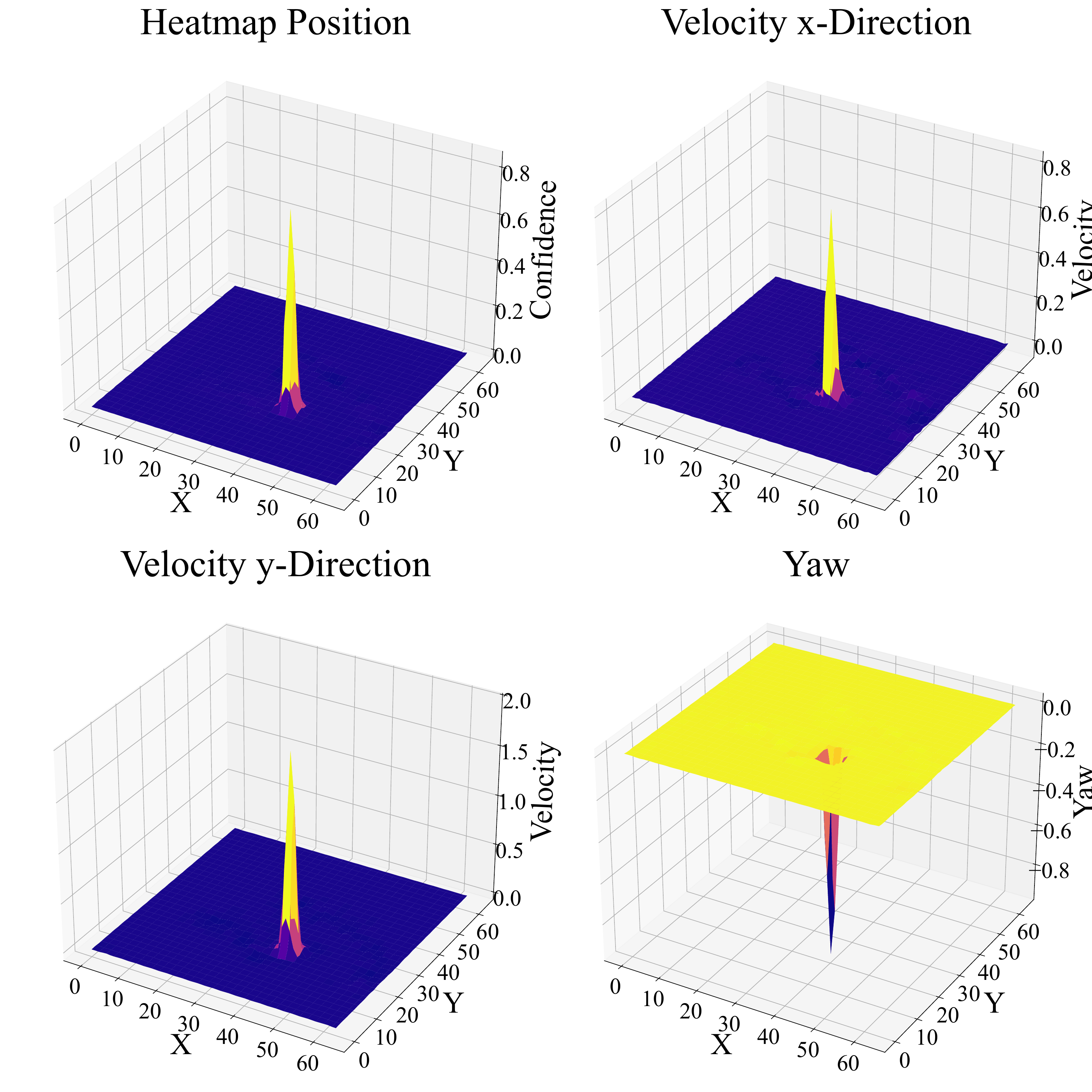}
    \caption{This figure shows an exemplary multi-channel output heatmap $\hat{\textbf{H}}$ consisting of channels for the position, velocity in both x- and y-direction, and the estimated yaw angle.}
    \label{fig:example:output}
\end{figure}

The architecture of \emph{TinyCenterSpeed} is shown in \Cref{fig:CS-ARCH}. The model is built in an hourglass structure, where the input features are compressed into a lower-dimensional feature map which represents the necessary information for the dense output predictions. The model predicts a multi-channel heatmap of size $k \times k$ with four channels $\hat{\textbf{H}} \in \mathbf{R}^{k\times k \times 4}$. A single channel in the heatmap is represented with $\hat{H} \in \mathbb{R}^{k\times k \times 1}$ and single points in the keymap are addressed as $\hat{h}_{i,\,j} \in \mathbb{R}$ with $i,\,j \in [1,\,\hdots ,\,k]$.
The first channel is interpreted as a two-dimensional spatial probability distribution of an object's position. The remaining three channels represent the dense predictions of the velocity and the orientation corresponding to the positions in the first channel. An exemplary output heatmap $\hat{\textbf{H}}$ is shown in \Cref{fig:example:output}.
The symbols $\textbf{H},\,H,\,h_{i,\,j}$ are then used to represent heatmap ground truth values. 

\subsection{Dataset Creation}
A dataset was developed to address the unique requirements of the specific environment under study (autonomous racing with 2D \gls{lidar}), as no pre-existing datasets adequately captured the characteristics or conditions necessary for this work. This dataset was specifically tailored to facilitate robust training and ensure the applicability of the methods within the intended context. Data was collected during the standard physical operation of two robots. While not as precise as most motion capturing methods, the proposed method allows for more variety and realism in the training distribution. The dataset is comprised of \gls{lidar} data $d_{lidar}$ from one robot (the \emph{ego} robot), and the \gls{gt} opponent state taken from onboard state estimation of the other robot (the \emph{opp} robot).
\begin{equation}
    \mathbf{D} = [d_{lidar},\mathbf{s_{opp}}]
\end{equation}
\begin{equation}
    \mathbf{s_{opp}}= [x_{opp},y_{opp},v_{x,opp},v_{y,opp}, \theta_{opp}]
\end{equation}
The static state $\mathbf{s_{ego}} = [x_{ego}, y_{ego},\theta_{ego}]$ of the robot providing $d_{lidar}$ is also obtained through the onboard state estimation.
The global \gls{gt} opponent state is transformed into a local \gls{gt} state with respect to $\mathbf{s_{ego}}$, removing the need to store transformations and enabling efficient preprocessing.
The global position of the \emph{ego} robot is then indicated with $x_{ego},\,y_{ego}$, the global attitude with $\theta_{ego}$ and similarly the local position, velocities, and attitude of the \emph{opp} robot in the \emph{ego} robot frame are indicated with $x_{opp},y_{opp},v_{x,opp},v_{y,opp}, \theta_{opp}$.
 The \gls{gt} from the state estimation was complemented with an accuracy-focused implementation of an \gls{abd}, which removes \gls{lidar} point cloud downsampling compared to the method proposed in \cite{baumann2024forzaeth} and performs offline detection. Together these \gls{gt} signals were used to filter and manually correct outliers in the dataset, enabling both easy data collection and high data quality.  For storing efficiency the dataset stores sparse  \gls{lidar} data which is preprocessed during training time.

\begin{figure}
    \centering
    \begin{subfigure}[b]{\linewidth}
        \centering
        \includegraphics[width=0.45\linewidth, trim=0cm 0cm 0cm 0.4cm, clip]{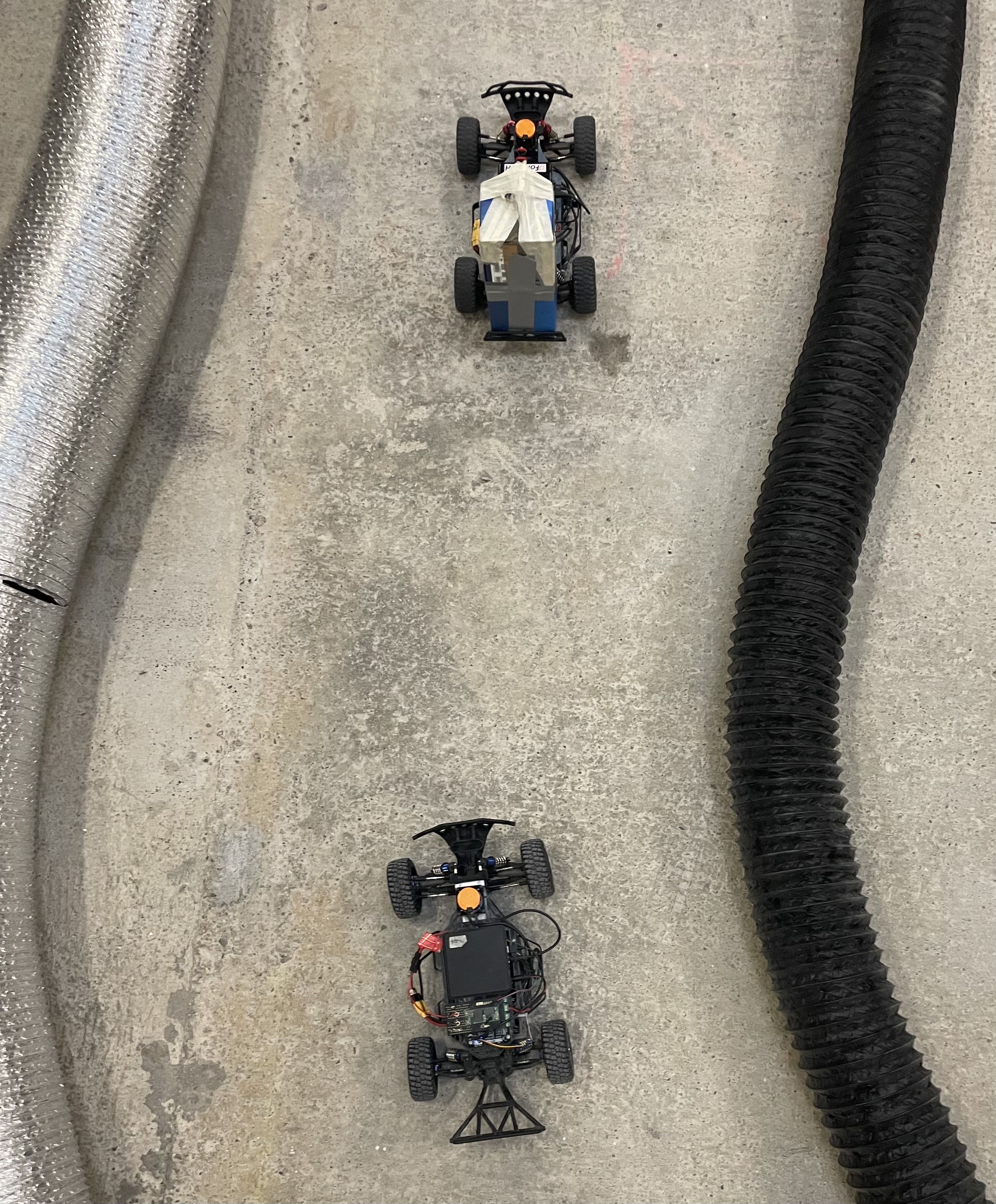}
        \caption{}
        \label{subfig:physical_cars}
    \end{subfigure}
    \hfill
    \centering
    \begin{subfigure}[b]{\linewidth}
        \centering
        \includegraphics[width=\linewidth]{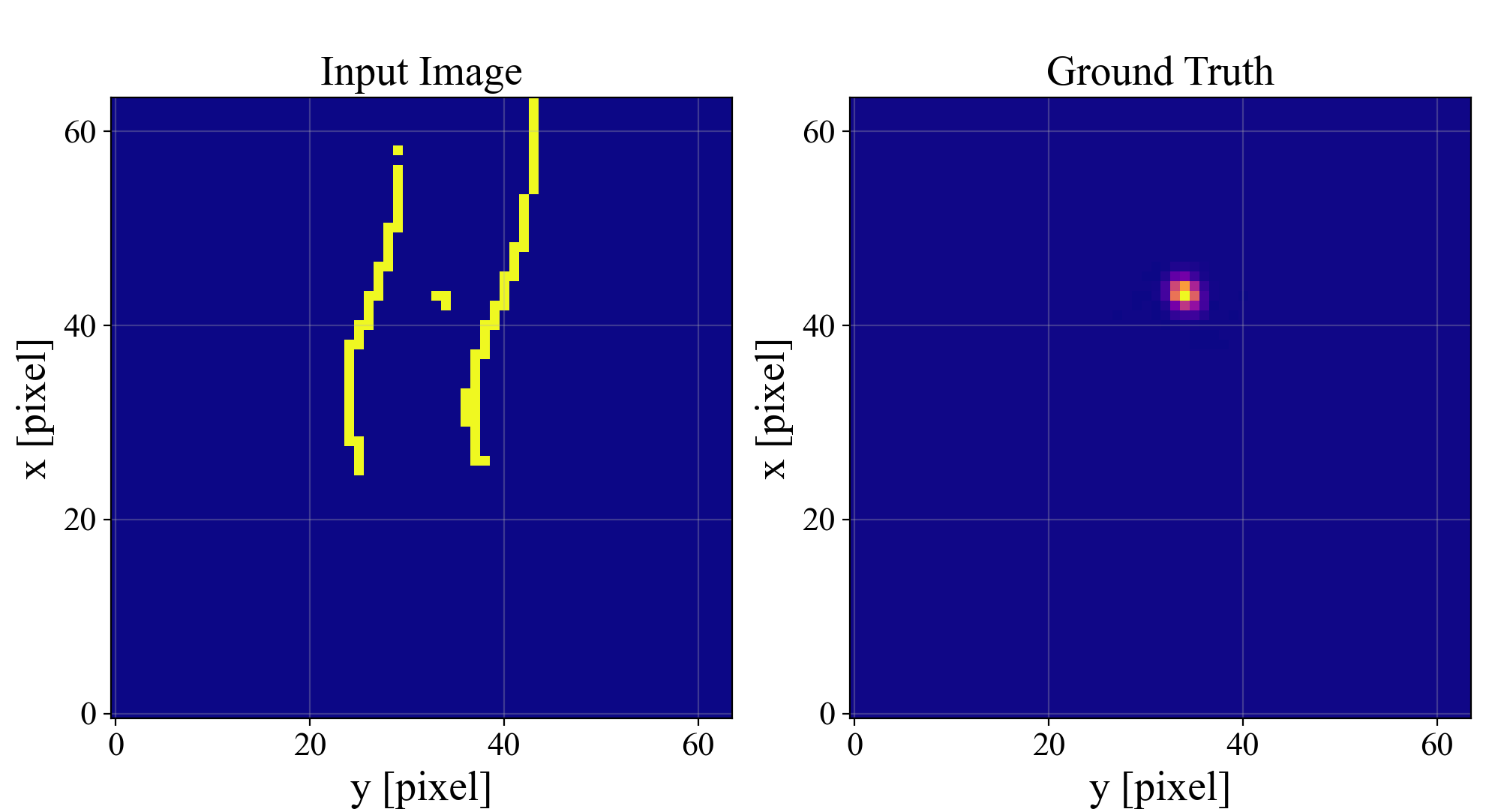} 
        \caption{}
        \label{subfig:data}
    \end{subfigure}
    \caption{The physical ego car behind the opponent car is shown here in \Cref{subfig:physical_cars}. The \gls{bev} preprocessed input and the corresponding positional ground truth are instead shown in \Cref{subfig:data}.}
    \label{fig:combined_data}
\end{figure}

During preprocessing $d_{lidar}$ is transformed into an image $\hat{\textbf{I}} \in \mathbf{R}^{k \times k \times 3}$ of fixed size $k \times k$ in \gls{bev} \cite{li2024delving,chen2017multi, zhou2018voxelnet, lang2019pointpillars}. The sparse polar coordinates of the \gls{lidar} scan are discretized using a fixed pixel size $p$. The image size $k$ and the pixel size $p$ determine the \gls{fov} $FOV = (k \times p)^2$ and the captured detail of the model input. One input frame $\hat{\textbf{I}} \in \mathbf{R}^{k \times k \times 3}$ consits of three explicitly defined features \cite{chen2017multi}. $\hat{\textbf{I}}$ contains a binary occupancy grid, the normalized intensity of the scans as well as the density of scan points per pixel. To obtain temporal information, two of these frames are concatenated as input for the model. Following \cite{zhou2019objects}, the dense \gls{gt} images are generated by drawing the \gls{gt} keypoints onto an image using a gaussian kernel $\hat{\textbf{O}}_{GT} = A \times \exp(-\frac{(x-x_{opp}) + (y - y_{opp})}{2\sigma_{GT}})$ where $A \in \{1, v_{x,opp}, v_{y,opp}, \theta_{opp}\}$ for the position, velocity and the orientation heatmaps respectively.
A combined representation showing the physical robot, a discretized occupancy grid in \gls{bev}, and the \gls{gt} position target is available in \Cref{fig:combined_data}.

The final positional and velocity output of the detection system is reported in the Frenet coordinate system, which utilizes a reference line to express position and velocity and as such can be helpful for downstream applications \cite{baumann2024PSpliner, MUZZINI2024gpuFrenet}. 
The coordinates are $s$ for longitudinal position on the reference line and $d$ for perpendicular lateral deviation from the reference line. Furthermore, $(v_s,\,v_d)$ respectively indicate the projection of the cartesian velocity $(v_x,\,v_y)$ on the tangent and perpendicular line to the point corresponding to $s$ on the reference line.
For more information and a graphical example, see \cite{baumann2024forzaeth}. Detailed information about the dataset can be found on \href{https://github.com/ForzaETH/TinyCenterSpeed/blob/971265a83b262f2d2fd781bc82ab146bf702c405/dataset/README.md}{Github}.

\subsection{Neural Network Training}
During training, several augmentation techniques are applied to increase variability in the training data and enhance generalization abilities. With probability $p = 0.5$, the input image is flipped along the x-axis and is rotated randomly with an angle $\phi \in [- \frac{\pi}{4}, \frac{\pi}{4}]$. Both the limitations on the values of $\phi$ and only considering flips along the x-axis come from the fact that the perceiving robot always sits at the coordinate origin and faces along the positive x-axis. To increase variability in the absolute values of the velocity data, a frame is skipped with probability $p = 0.5$ when constructing the model input. In that case, the \gls{gt} velocity is artificially doubled, significantly increasing the value range represented in the dataset. 
An ablation to assess the significance of these augmentation is later performed in \Cref{subsec:data_aug}.

Training the dense predictions introduces some challenges. Since most \gls{gt} heatmaps are sparse in the sense that only a small proportion of the pixels should indeed hold a value greater than zero, models often tend to ignore all inputs. To counteract this, \emph{TinyCenterSpeed} was trained using a variation of the \emph{Heatmap Weighted Loss} \cite{heatmaploss}.

\begin{equation}
    \resizebox{0.9\linewidth}{!}{$
    \begin{aligned}
        \mathcal{L}(\textbf{H}, \hat{\textbf{H}}) &= \alpha \mathcal{L}_{hm}(H_{pos},\hat{H}_{pos}) \\
        &\quad + (1-\alpha)(\mathcal{L}_{hm}(H_{v}, \hat{H}_{v})+\mathcal{L}_{hm}(H_{yaw}, \hat{H}_{yaw})) \\
    \end{aligned}
    $}
    \label{eq:total-loss}
\end{equation}

\begin{equation}
    \mathcal{L}_{hm}(H,\hat{H}) = \sum_{i=0}^k\sum_{j=0}^k(1+h_{i,j})(h_{i,j}-\hat{h}_{i,j})^2
    \label{eq:hm-loss}
\end{equation}

The total loss $\mathcal{L}(\textbf{H}, \hat{\textbf{H}})$ results as a weighted sum of heatmap losses $\mathcal{L}_{hm}(H, \hat{H})$, which, taken singularly as in \Cref{eq:hm-loss}, reflect the original implementation of \cite{heatmaploss} used to increase the significance of non-zero values at keypoint locations. The parameter $\alpha$ weighs the loss of the position heatmap $H_{pos}$ against the combined loss of the velocity $H_{v}$ and orientation heatmaps $H_{yaw}$. This is used to control the asymmetric importance of knowledge about position versus knowledge about an object's velocity and orientation.

\subsection{Model Details}
\emph{TinyCenterSpeed} uses two $3 \times 3$ convolutional layers with stride 2 followed by 2D-batchnorm and ReLU activation function. The multi-channel heatmaps are predicted using two $3 \times 3$ transposed convolutions with stride 2 and ReLU. For the main experiments, an image size $k = 64$ and a pixel size $p = 0.1$ were used. The model was optimized using Adam \cite{kingma2017adammethodstochasticoptimization} with a learning rate of $5\mathrm{e}{-}5$
. During training, $\alpha = 0.99$ and a batch-size of 32 were used.

\subsection{Kalman Filter Tracker}
\emph{TinyCenterSpeed} is designed to detect the position and estimate the velocity of multiple objects directly from sequential frames. This method operates effectively without the need for explicit considerations of physical constraints and temporality. Although \emph{TinyCenterSpeed} can provide velocity estimates independently, integrating a Kalman Filter can still be beneficial to refine these raw measurements by filtering and smoothing the outputs. In contrast, the \gls{abd} method requires the addition of a Kalman Filter to derive velocity estimates. Within this framework, when a tracker is mentioned in the results section, it specifically refers to the use of a Kalman Filter, as implemented in \cite{baumann2024forzaeth}.

\subsection{Model Deployment}
\begin{figure}[!htb]
    \centering
    \includegraphics[width=0.8\columnwidth, trim=0cm 1cm 0cm 0cm, clip]{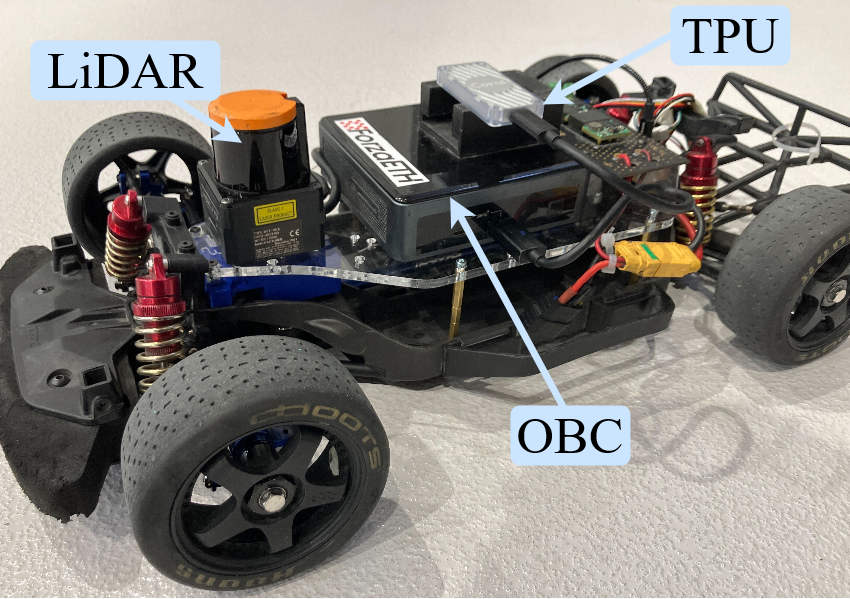}
    \caption{The \emph{ForzaETH} racecar with the \gls{lidar} sensor, the \gls{obc} and the external \emph{Google Coral} \gls{tpu} connected via USB. This allows offloading of the \gls{nn} inference from the main \gls{obc} to the \gls{tpu} when the \gls{nn} is \texttt{INT8} quantized.}
    \label{fig:tpu}
\end{figure}

To ensure computational efficiency during inference, \emph{TinyCenterSpeed} can be \texttt{INT8} quantized and deployed on a \gls{tpu} (\emph{Google Coral}, shown in \Cref{fig:tpu}), allowing the system to maintain high performance while reducing its computational overhead such that the \gls{obc} can dedicate more \gls{cpu} utilization to other high-intensity autonomy computations such as \gls{slam}, \gls{mpc}, or planning. For full integer quantization, the \texttt{PyTorch} trained model is exported via \texttt{ONNX} to \texttt{tflite} to then perform the inference via the \texttt{PyCoral} API\footnote{https://coral.ai/docs/reference/py/}.

\begin{table*}[!htb] 
\small
    \centering 
    \setlength{\tabcolsep}{0.7mm}
    \begin{adjustbox}{max width=\textwidth}
    \begin{tabular}{l|c|c|cc|cc|cc|cc|c}
    \toprule
    \textbf{Perception Method} & \textbf{Quant} & \textbf{Tracking} & \multicolumn{2}{c|}{\textbf{\bm{$s$} [m] $\downarrow$}} & \multicolumn{2}{c|}{\textbf{\bm{$d$} [m]} $\downarrow$} & \multicolumn{2}{c|}{\textbf{\bm{$v_s$} [m/s]} $\downarrow$} & \multicolumn{2}{c|}{\textbf{\bm{$v_d$} [m/s]} $\downarrow$} & \textbf{Improve [\%] $\bm{\uparrow}$} \\
    \textbf{(Single Opponent)} &  & & $\bm{\mu_s}$ & $\bm{\sigma_s}$ & $\bm{\mu_d}$ & $\bm{\sigma_d}$ & $\bm{\mu_{v_s}}$ & $\bm{\sigma_{v_s}}$ & $\bm{\mu_{v_d}}$ & $\bm{\sigma_{v_d}}$ \\
    \midrule
    ABD & - & \xmark & 0.21 & 0.15 & 0.10 & 0.09 & - & - & - & - & - \\
    TinyCenterSpeed & \xmark & \xmark & \textbf{0.07} &\textbf{0.04} & \textbf{0.06} & \textbf{0.05} & \textbf{0.38}  & \textbf{0.37}  & \textbf{0.37}  & \textbf{0.36}  & \textbf{58.76} \\
    TinyCenterSpeed & \checkmark & \xmark & \textbf{0.07} & \textbf{0.04} & 0.07 & 0.06 & 0.46  & 0.43  & 0.40 & 0.39  & 48.92 \\
    \midrule
     ABD & - & \checkmark & 0.19 & 0.14 & 0.08 & 0.07 & 1.06 & 1.06 & 0.16 & 0.16 & - \\
    TinyCenterSpeed & \xmark & \checkmark & \textbf{0.10} & \textbf{0.05} & \textbf{0.02} & \textbf{0.02} & \textbf{0.22} & \textbf{0.16} & \textbf{0.10} & \textbf{0.10} & \textbf{61.38} \\
    TinyCenterSpeed & \checkmark & \checkmark & 0.11 & \textbf{0.05} & 0.03 & \textbf{0.02} & 0.30 & 0.18 & \textbf{0.10} & \textbf{0.10} & 58.76 \\
    \bottomrule
    \end{tabular}
    \end{adjustbox}
    \caption{Quantitative Comparison of Perception Methods for a single opponent. The metrics reported include the \gls{rmse} ($\bm{\mu}$) and standard deviation of the absolute error ($\bm{\sigma}$) for Frenet longitudinal position ($s$ [m]), Frenet lateral position ($d$ [m]), Frenet longitudinal speed  ($v_s$ [m/s]), and Frenet lateral speed ($v_d$ [m/s]). Average \gls{rmse} improvement is further reported ([\%]). Performance is measured both with and without tracking and the usage of quantization to enable \gls{tpu} inference.}
    \label{tab:err}
\end{table*}

\section{Results}
\emph{TinyCenterSpeed} is evaluated against the current, race-winning perception system of \emph{ForzaETH}'s race stack \cite{baumann2024forzaeth} which uses an \gls{abd}. To assess perception quality the \gls{rmse} is used.
\begin{equation}
    \text{RMSE} = \sqrt{\frac{1}{N} \sum_{i=1}^N \left( y_i - \hat{y}_i \right)^2}
\end{equation}
The perception systems are evaluated on identical input data recorded in the form of a \texttt{rosbag}.

\subsection{Single Opponent Evaluation}

In the context of single opponent tracking, the \gls{abd}, as detailed in \cite{baumann2024forzaeth}, is designed specifically for a single adversary. As indicated in \Cref{tab:err}, the identical metrics are applied to assess performance relative to the \emph{TinyCenterSpeed} system. Initially, positional accuracy is evaluated using the \gls{rmse} for $s$ and $d$ coordinates. Subsequently, velocity states are assessed with the usage of the tracking \gls{ekf} from \cite{baumann2024forzaeth}, focusing on the \gls{rmse} values for longitudinal and lateral velocities. The tracking feature is necessary for the \gls{abd} to enable velocity estimation. In contrast, \emph{TinyCenterSpeed} is capable of directly generating velocity measurements without the \gls{ekf}, however it is still capable of utilizing the tracker for enhanced accuracy.

The comparative results highlight that \emph{TinyCenterSpeed} significantly enhances the detection of a single opponent across all evaluated metrics. Specifically, without tracker utilization, \emph{TinyCenterSpeed} enhances average detection performance by 58.76\%. When deploying the quantized model to leverage \gls{tpu} inference capabilities, there is a slight reduction in performance, yet it still yields a 48.92\% overall improvement in both positional and velocity states.

Integrating the tracker not only aids the \gls{abd} in velocity estimation but also refines \emph{TinyCenterSpeed}'s detection capabilities through the \gls{ekf}'s filtering and smoothing effects. With the tracker, \emph{TinyCenterSpeed} achieves an average performance enhancement of 61.38\%. However, upon offloading on the \gls{tpu}, a minor decrement to 58.76\% is observed, attributed to the quantization.

\subsection{Multiple Opponents Evaluation}
In contrast to the \gls{abd} system outlined in \cite{baumann2024forzaeth}, the \emph{TinyCenterSpeed} framework demonstrates superior capabilities in managing multiple opponents simultaneously without relying on the \gls{ekf} tracker. The efficacy of this approach is substantiated in \Cref{tab:multiopp} utilizing the \gls{mate} and \gls{mave} metrics, which were established and popularised in \cite{caesar2020nuscenes}. In this experiment three cars were put on a track, hence \emph{TinyCenterSpeed} had to estimate both position and velocity of two opponents simultaneously.

The analysis reveals that positional errors average only \SI{0.16}{\metre}, while velocity discrepancies are at \SI{0.32}{\metre\per\second}. Notably, both the quantized and full-precision versions of the network exhibit very similar performance, underscoring \emph{TinyCenterSpeed}'s robustness not only against the previous \gls{sota} in autonomous racing perception, namely \gls{abd} for single opponents, but also in scenarios involving multiple adversaries.

For comparative insight, it is noteworthy that full-scale autonomous driving systems, which employ extensive \gls{nn} architectures and virtually no computational restrictions, achieve comparable \gls{mate} and \gls{mave} values on the \emph{nuScenes} dataset. For instance, \emph{CenterPoint} \cite{yin2021center} reports a \gls{mate} of \SI{0.249}{\metre} and a \gls{mave} of \SI{0.250}{\metre\per\second}, while \emph{BevDet} \cite{huang2021bevdet} records a \gls{mate} of \SI{0.529}{\metre} and a \gls{mave} of \SI{0.979}{\metre\per\second}. Although these systems operate on vastly different datasets, making direct a comparison intractable, it is still interesting to mention.

\begin{table}[!htb] 
\small
    \centering 
    \setlength{\tabcolsep}{0.7mm}
    \begin{tabular}{l|c|c|c}
    \toprule
    \textbf{Perception Method} & \textbf{Quant} & \textbf{mATE [m]$\downarrow$} & \textbf{mAVE [m/s]$\downarrow$} \\
    \midrule
    TinyCenterSpeed & \xmark & 0.159 & 0.318 \\
    TinyCenterSpeed & \checkmark & 0.159 & 0.321 \\
    \bottomrule
    \end{tabular}
    \caption{Quantitative Comparison of Perception Methods for multiple opponents. The multi-opponent metrics reported include the \gls{mate} and \gls{mave}. Performance is measured with the usage of quantization to enable \gls{tpu} inference.}
    \label{tab:multiopp}
\end{table}

\subsection{The Effect of Data Augmentation} \label{subsec:data_aug}
\emph{TinyCenterSpeed} incorporates various data augmentation techniques during training. These include randomly rotating samples around the coordinate origin (\emph{Rot.}), flipping them along the x-axis (\emph{Flip}), and artificially doubling the observed speed by skipping one frame (\emph{Speed}). As shown in \cref{tab:ablation:augmentation}, applying all these techniques simultaneously produces the best results, with the additional velocity augmentation significantly enhancing velocity estimation.

\begin{table}[!htb] 
\small
    \centering 
    \setlength{\tabcolsep}{0.7mm}
    \begin{tabular}{ccc|cccc}
    \toprule
    \multicolumn{3}{c|}{\textbf{Augmentation}} & \multicolumn{4}{c}{\textbf{RMSE}} \\
    \midrule
    Rot. & Flip & Speed & $s$ [m] $\downarrow$ & $d$ [m] $\downarrow$ & $v_s$ [m/s] $\downarrow$ & $v_d$ [m/s] $\downarrow$ \\

    \midrule
    \xmark & \xmark & \xmark & 0.09 & 0.07 & 1.00 & 0.37 \\
    \checkmark & \checkmark & \xmark & 0.08 & 0.06 & 0.92 & \textbf{0.27} \\
    \checkmark & \checkmark & \checkmark & \textbf{0.07} & \textbf{0.06} & \textbf{0.38} & 0.33 \\
    \bottomrule
    \end{tabular}
    \caption{Effect of data augmentation techniques on evaluation metrics. \emph{Rot.}: the sample is randomly rotated. \emph{Flip}: the sample is randomly flipped along the x-axis. \emph{Speed}: the observed velocity is artificially doubled by skipping a frame.}
    \label{tab:ablation:augmentation}
\end{table}

\subsection{Latency and CPU Utilization}
As in \Cref{tab:results:latency}, the latency was evaluated using the \gls{obc} of the \emph{ForzaETH} racecar (Intel NUC: Intel Core i7-10710, 6-Core, 1.1GHz) with and without an external \gls{tpu} (Google Coral USB TPU Accelerator). The latency of \emph{TinyCenterSpeed} with and without additional filtering was evaluated to be $8.57ms$ and $8.56ms$ on average,  respectively. The CPU utilization, measured in utilization of CPU threads, was increased from $73.07\%$ (ABD) to $387.3\%$. \emph{TinyCenterSpeed} can be compiled and deployed on an external \gls{tpu}, which decreases both the latency to $7.52ms$ and the \gls{cpu} utilization to $50.26\%$. This results in a factor 8.3$\times$ reduction of computational utilization on the \gls{obc}.

\begin{table}[!htb]
    \centering
    \begin{adjustbox}{max width=\columnwidth}
    \begin{tabular}{l|c|c|ccc}
    \toprule
    \textbf{Method} & \textbf{Quant} & \textbf{Tracking} & \textbf{$\bm{\mu_t}$ [ms] $\downarrow$} & \textbf{$\bm{\sigma_t}$ [ms] $\downarrow$} & \textbf{CPU [\%] $\downarrow$} \\
    \midrule
    ABD                       & -              & \xmark            & 10.55 & 1.04 & 52.83 \\
    ABD                       & -              & \checkmark        & 12.87 & 1.73 & 73.07 \\
    TinyCenterSpeed           & \xmark         & \xmark            & 8.56 & 9.96& 377.10\\
    TinyCenterSpeed           & \xmark         & \checkmark        & 8.57 & 6.74 & 387.30\\
    TinyCenterSpeed           & \checkmark     & \xmark            & \textbf{7.88}& \textbf{0.42} & \textbf{45.34} \\
    TinyCenterSpeed           & \checkmark     & \checkmark        & 8.27 & 0.73 & 53.26\\
    \bottomrule
    \end{tabular}
    \end{adjustbox}
    \caption{Showing the mean and standard deviation of the latency ($\mu_t$, $\sigma_t$) as well as the average \gls{cpu} utilization during inference. Note that 100\% utilitation corresponds to full utilization of a single \gls{cpu} thread. The values were obtained on an Intel NUC (Intel Core i7-10710, 6-Core, 1.1GHz) and quantized models utilized a \gls{tpu} (Google Coral).}
    \label{tab:results:latency}
\end{table}

\section{Conclusion}
In this study, \emph{TinyCenterSpeed} is introduced as an \gls{nn}-based model for opponent estimation in scaled autonomous racing. The full-scale and general autonomous driving \gls{nn} architecture of \emph{CenterPoint} is adapted to meet the computational constraints of 1:10 scale autonomous racing cars. To the best of our knowledge, this represents the first \gls{nn}-based implementation of scaled autonomous detection and velocity-estimation using 2D \gls{lidar} in this context.
An improvement of up to 61.38\% in spatial and velocity estimation is achieved within the autonomous racing domain by \emph{TinyCenterSpeed}, which also facilitates the tracking of multiple opponents without the necessity for a \gls{ekf} tracker, though it retains compatibility with such systems. Additionally, model quantization is employed, allowing the offloading of detection tasks to an external on-board \gls{tpu}. This strategy results in an 8.3-fold reduction in \gls{cpu} usage, thereby freeing the main \gls{obc} to manage other autonomous computations. The \gls{nn} inference on the \gls{tpu} is completed in only \SI{7.88}{\milli\second}.

\bibliographystyle{IEEEtran}
\bibliography{main}
\end{document}

%% file: glossaries.tex
\usepackage[acronym]{glossaries}
\makeglossaries

\newacronym{ml}{ML}{Machine Learning}
\newacronym{bev}{BEV}{Bird's-eye-view}
\newacronym{gt}{GT}{Ground-Truth}
\newacronym{lidar}{LiDAR}{Light Detection and Ranging}
\newacronym{abd}{ABD}{Adaptive Breakpoint Detector}
\newacronym{rmse}{RMSE}{Root Mean Squared Error}
\newacronym{tpu}{TPU}{Tensor Processing Unit}
\newacronym{fov}{FOV}{Field of View}
\newacronym{sota}{SotA}{State-of-the-Art}
\newacronym{kf}{KF}{Kalman Filter}
\newacronym{nn}{NN}{Neural Network}
\newacronym{gpu}{GPU}{Graphical Processing Unit}
\newacronym{cpu}{CPU}{Central Processing Unit}
\newacronym{mave}{mAVE}{mean Average Velocity Error}
\newacronym{mate}{mATE}{mean Average Translation Error}
\newacronym{ekf}{EKF}{Extended Kalman Filter}
\newacronym{obc}{OBC}{On-Board Computer}
\newacronym{slam}{SLAM}{Simultaneous Localization And Mapping}
\newacronym{mpc}{MPC}{Model Predictive Control}
\newacronym{gui}{GUI}{Graphical User Interface}

%% file: root.bbl
\begin{thebibliography}{10}
\providecommand{\url}[1]{#1}
\csname url@samestyle\endcsname
\providecommand{\newblock}{\relax}
\providecommand{\bibinfo}[2]{#2}
\providecommand{\BIBentrySTDinterwordspacing}{\spaceskip=0pt\relax}
\providecommand{\BIBentryALTinterwordstretchfactor}{4}
\providecommand{\BIBentryALTinterwordspacing}{\spaceskip=\fontdimen2\font plus
\BIBentryALTinterwordstretchfactor\fontdimen3\font minus \fontdimen4\font\relax}
\providecommand{\BIBforeignlanguage}[2]{{%
\expandafter\ifx\csname l@#1\endcsname\relax
\typeout{** WARNING: IEEEtran.bst: No hyphenation pattern has been}%
\typeout{** loaded for the language `#1'. Using the pattern for}%
\typeout{** the default language instead.}%
\else
\language=\csname l@#1\endcsname
\fi
#2}}
\providecommand{\BIBdecl}{\relax}
\BIBdecl

\bibitem{caesar2020nuscenes}
H.~Caesar, V.~Bankiti, A.~H. Lang, S.~Vora, V.~E. Liong, Q.~Xu, A.~Krishnan, Y.~Pan, G.~Baldan, and O.~Beijbom, ``nuscenes: A multimodal dataset for autonomous driving,'' in \emph{Proceedings of the IEEE/CVF conference on computer vision and pattern recognition}, 2020, pp. 11\,621--11\,631.

\bibitem{waymo}
P.~Sun, H.~Kretzschmar, X.~Dotiwalla, A.~Chouard, V.~Patnaik, P.~Tsui, J.~Guo, Y.~Zhou, Y.~Chai, B.~Caine \emph{et~al.}, ``Scalability in perception for autonomous driving: Waymo open dataset,'' in \emph{Proceedings of the IEEE/CVF conference on computer vision and pattern recognition}, 2020, pp. 2446--2454.

\bibitem{kitti}
A.~Geiger, P.~Lenz, C.~Stiller, and R.~Urtasun, ``Vision meets robotics: The kitti dataset,'' \emph{The International Journal of Robotics Research}, vol.~32, no.~11, pp. 1231--1237, 2013.

\bibitem{yin2021center}
T.~Yin, X.~Zhou, and P.~Krähenbühl, ``Center-based 3d object detection and tracking,'' in \emph{2021 IEEE/CVF Conference on Computer Vision and Pattern Recognition (CVPR)}, 2021, pp. 11\,779--11\,788.

\bibitem{okelly2019f1tenth}
\BIBentryALTinterwordspacing
M.~O'Kelly, H.~Zheng, D.~Karthik, and R.~Mangharam, ``F1tenth: An open-source evaluation environment for continuous control and reinforcement learning,'' in \emph{Proceedings of the NeurIPS 2019 Competition and Demonstration Track}, ser. Proceedings of Machine Learning Research, H.~J. Escalante and R.~Hadsell, Eds., vol. 123.\hskip 1em plus 0.5em minus 0.4em\relax PMLR, 08--14 Dec 2020, pp. 77--89. [Online]. Available: \url{https://proceedings.mlr.press/v123/o-kelly20a.html}
\BIBentrySTDinterwordspacing

\bibitem{baumann2024forzaeth}
N.~Baumann, E.~Ghignone, J.~K{\"u}hne, N.~Bastuck, J.~Becker, N.~Imholz, T.~Kr{\"a}nzlin, T.~Y. Lim, M.~L{\"o}tscher, L.~Schwarzenbach \emph{et~al.}, ``Forzaeth race stack—scaled autonomous head-to-head racing on fully commercial off-the-shelf hardware,'' \emph{Journal of Field Robotics}, 2024.

\bibitem{cai2023bevfusion4d}
\BIBentryALTinterwordspacing
H.~Cai, Z.~Zhang, Z.~Zhou, Z.~Li, W.~Ding, and J.~Zhao, ``Bevfusion4d: Learning lidar-camera fusion under bird's-eye-view via cross-modality guidance and temporal aggregation,'' 2023. [Online]. Available: \url{https://arxiv.org/abs/2303.17099}
\BIBentrySTDinterwordspacing

\bibitem{chen2021pseudoimage}
G.~Chen, F.~Wang, S.~Qu, K.~Chen, J.~Yu, X.~Liu, L.~Xiong, and A.~Knoll, ``Pseudo-image and sparse points: Vehicle detection with 2d lidar revisited by deep learning-based methods,'' \emph{IEEE Transactions on Intelligent Transportation Systems}, vol.~22, no.~12, pp. 7699--7711, 2021.

\bibitem{jia2022twoDvsthreeD}
D.~Jia, A.~Hermans, and B.~Leibe, ``2d vs. 3d lidar-based person detection on mobile robots,'' in \emph{2022 IEEE/RSJ International Conference on Intelligent Robots and Systems (IROS)}, 2022, pp. 3604--3611.

\bibitem{baumann2024PSpliner}
N.~Baumann, E.~Ghignone, C.~Hu, B.~Hildisch, T.~Hämmerle, A.~Bettoni, A.~Carron, L.~Xie, and M.~Magno, ``Predictive spliner: Data-driven overtaking in autonomous racing using opponent trajectory prediction,'' \emph{IEEE Robotics and Automation Letters}, pp. 1--8, 2024.

\bibitem{hell2024lidar}
M.~Hell, G.~Hajgato, A.~Bogar-Nemeth, and G.~Bari, ``A lidar-based approach to autonomous racing with model-free reinforcement learning,'' in \emph{2024 IEEE Intelligent Vehicles Symposium (IV)}, 2024, pp. 258--263.

\bibitem{zarrar2024tinylidarnet}
M.~M. Zarrar, Q.~Weng, B.~Yerjan, A.~Soyyigit, and H.~Yun, ``Tinylidarnet: 2d lidar-based end-to-end deep learning model for f1tenth autonomous racing,'' in \emph{2024 IEEE/RSJ International Conference on Intelligent Robots and Systems (IROS)}, 2024, pp. 2878--2884.

\bibitem{zhou2018voxelnet}
Y.~Zhou and O.~Tuzel, ``Voxelnet: End-to-end learning for point cloud based 3d object detection,'' in \emph{Proceedings of the IEEE conference on computer vision and pattern recognition}, 2018, pp. 4490--4499.

\bibitem{lang2019pointpillars}
A.~H. Lang, S.~Vora, H.~Caesar, L.~Zhou, J.~Yang, and O.~Beijbom, ``Pointpillars: Fast encoders for object detection from point clouds,'' in \emph{Proceedings of the IEEE/CVF conference on computer vision and pattern recognition}, 2019, pp. 12\,697--12\,705.

\bibitem{li2024delving}
H.~Li, C.~Sima, J.~Dai, W.~Wang, L.~Lu, H.~Wang, J.~Zeng, Z.~Li, J.~Yang, H.~Deng, H.~Tian, E.~Xie, J.~Xie, L.~Chen, T.~Li, Y.~Li, Y.~Gao, X.~Jia, S.~Liu, J.~Shi, D.~Lin, and Y.~Qiao, ``Delving into the devils of bird’s-eye-view perception: A review, evaluation and recipe,'' \emph{IEEE Transactions on Pattern Analysis and Machine Intelligence}, vol.~46, no.~4, pp. 2151--2170, 2024.

\bibitem{zhao2024bev}
\BIBentryALTinterwordspacing
J.~Zhao, J.~Shi, and L.~Zhuo, ``Bev perception for autonomous driving: State of the art and future perspectives,'' \emph{Expert Systems with Applications}, vol. 258, p. 125103, 2024. [Online]. Available: \url{https://www.sciencedirect.com/science/article/pii/S0957417424019705}
\BIBentrySTDinterwordspacing

\bibitem{chen2017multi}
X.~Chen, H.~Ma, J.~Wan, B.~Li, and T.~Xia, ``Multi-view 3d object detection network for autonomous driving,'' in \emph{Proceedings of the IEEE conference on Computer Vision and Pattern Recognition}, 2017, pp. 1907--1915.

\bibitem{zhou2019objects}
X.~Zhou, D.~Wang, and P.~Kr{\"a}henb{\"u}hl, ``Objects as points,'' \emph{arXiv preprint arXiv:1904.07850}, 2019.

\bibitem{heatmaploss}
S.~Li and X.~Xiang, ``Lightweight human pose estimation using loss weighted by target heatmap,'' in \emph{International Conference on Pattern Recognition}.\hskip 1em plus 0.5em minus 0.4em\relax Springer, 2022, pp. 64--78.

\bibitem{MUZZINI2024gpuFrenet}
\BIBentryALTinterwordspacing
F.~Muzzini, N.~Capodieci, F.~Ramanzin, and P.~Burgio, ``Gpu implementation of the frenet path planner for embedded autonomous systems: A case study in the f1tenth scenario,'' \emph{Journal of Systems Architecture}, vol. 154, p. 103239, 2024. [Online]. Available: \url{https://www.sciencedirect.com/science/article/pii/S1383762124001760}
\BIBentrySTDinterwordspacing

\bibitem{kingma2017adammethodstochasticoptimization}
\BIBentryALTinterwordspacing
D.~P. Kingma and J.~Ba, ``Adam: A method for stochastic optimization,'' 2017. [Online]. Available: \url{https://arxiv.org/abs/1412.6980}
\BIBentrySTDinterwordspacing

\bibitem{huang2021bevdet}
J.~Huang, G.~Huang, Z.~Zhu, Y.~Ye, and D.~Du, ``Bevdet: High-performance multi-camera 3d object detection in bird-eye-view,'' \emph{arXiv preprint arXiv:2112.11790}, 2021.

\end{thebibliography}
